\documentclass[a4paper,11pt]{article}

\usepackage{fourier}
\usepackage{color}
\usepackage{graphicx}
\usepackage{url}
\usepackage[affil-it]{authblk}
\usepackage{amsmath}
\usepackage{wrapfig}
\usepackage{xspace}
\usepackage{floatrow}
\usepackage[T1]{fontenc}
\usepackage{times}
\usepackage{float}
\usepackage{booktabs}
\usepackage{multicol}
\usepackage{capt-of}
\usepackage{subcaption}
\usepackage[font={small}]{caption}

\graphicspath{{images/}}
\usepackage{color,soul}

\usepackage{comment}

\newfloatcommand{capbtabbox}{table}[][\FBwidth]

\begin{document}

\title{Open Source Dataset and Deep Learning Models for Online Digit Gesture Recognition on Touchscreens}

\author{Philip J. Corr, Guenole C. Silvestre and Chris J. Bleakley}
\affil{School of Computer Science, University College Dublin, Belfield, Dublin 4, Ireland.}
\date{}
\maketitle
\thispagestyle{empty}

\begin{abstract}
This paper presents an evaluation of deep neural networks for recognition of digits entered by users on a smartphone touchscreen. 
A new large dataset of Arabic numerals was collected for training and evaluation of the network. 
The dataset consists of spatial and temporal touch data recorded for 80 digits entered by 260 users.
Two neural network models were investigated.
The first model was a 2D convolutional neural (ConvNet) network applied to bitmaps of the glpyhs created by interpolation of the sensed screen touches and its topology is similar to that of previously published models for offline handwriting recognition from scanned images.
The second model used a 1D ConvNet architecture but was applied to the sequence of polar vectors connecting the touch points.
The models were found to provide accuracies of 98.50\% and 95.86\%, respectively.
The second model was much simpler, providing a reduction in the number of parameters from 1,663,370 to 287,690. 
The dataset has been made available to the community as an open source resource.
\end{abstract}

\section{Introduction}
Touchscreens are now pervasively used in smartphones and computing tablets.
Text input on a touchscreen commonly uses a virtual keyboard.
Unfortunately, the virtual keyboard occupies a significant portion of the screen.
This loss of screen is noticeable on smartphones but is especially problematic on smaller devices, such as smartwatches. 
Text entry by means of handwriting using the finger or thumb has the advantage that the gestures can be performed on top of a screen image or background.
Smaller screens can be easily accommodated by entering characters individually, one top of another \cite{kienzle2013writing}.

Previous work on handwriting recognition has mainly focused on processing images of pen-on-paper writing, i.e. offline character recognition.
Notably, the MNIST dataset was created using images of handwritten US census returns \cite{MNIST-dataset-98}.
Excellent recognition accuracy (99.2\%) was demonstrated on the MNIST dataset using a convolutional neural network (ConvNet) \cite{YannLecun-98}.
In contrast, online character recognition systems take input in the form of the continuously sensed position of the pen, finger, or thumb.
Online systems have the advantage of recording temporal information as well as spatial information.
To date, most work on online character recognition has focused on pen based systems \cite{LeCun-touch-terminal,LeRec,Verma-2004,Bahlmann}.
LeCun et al.'s paper proposed a ConvNet approach to the problem, achieving 96\% accuracy.
The method involved considerable preprocessing without which accuracy falls to 60\%.
The preprocessing step requires that the entire glyph is known \textit{a priori}, removing the possibility of early recognition and completion of the glyph. 

To date, there has been almost no work on using neural networks for online recognition of touchscreen handwriting using a finger or thumb.
Our observation is that digits formed using a finger or thumb have greater variability than those formed using a pen, with more examples of poorly formed glyphs.
Most likely, this is due to the users having better fine grained control of the pen. 
Furthermore, to enable operation on low cost, small form factor devices it is desirable that the resource footprint of the recognizer is low in terms of computational complexity and memory requirements.
To date, an unexplored dimension of the problem is that online entry allows early recognition and confirmation of the character entered, enabling faster text entry.

Herein, we report on a investigation seeking to address these challenges.
A large dataset of Arabic numerals was collected using a smartphone.
A number of deep learning models were explored and their accuracy evaluated for the collected dataset.
Of these architectures, two are reported herein.
The first model uses an approach similar to offline character recognition systems, i.e. a 2D ConvNet taking the bitmap of the completed glyph as input.
The second model uses a 1D ConvNet applied to the polar vector connecting touch positions.
The accuracy and the size of the networks are reported herein together with an analysis of some of the errors.
In addition, initial results on early digit recognition are provided.
To the best of our knowledge, this is the first work to report on a low footprint recognizer using polar vector inputs for online finger or thumb touch digit recognition.

\section{Dataset}

A software application was developed to record the dataset.
Prior to participation, subjects signed a consent form.
The application firstly asked subjects to enter their age, sex, nationality and handedness. 
Each subject was then instructed to gesture digits on the touchscreen using their index finger.
The digits 0 to 9 were entered four times. 
The sequence of digit entry was random. 
Instructions to the user were provided using voice synthesis to avoid suggesting a specific glyph rendering. 
The process was repeated for input using the thumb while holding the device with the same hand. 
This is to allow for applications where the user may only have one hand free. 
Cubic interpolation of touches during gesture input was rendered on the screen to provide visual feedback to the subject and to compute arclengths.
The screen was initially blank (white) and the gestures were displayed in black. 
The subject could use most of screen to gesture with small areas at the top and bottom reserved for instructions/interactions/guidance.
The subject was permitted to erase and repeat the entry, if desired. 


The dataset was acquired on a 4.7 inch iPhone 6 running iOS 10. 
Force touch data was not available. 
The touch panel characteristics are not publicly available, specifically the sampling frequency and spatial accuracy are unknown.
Values of 60Hz and $\pm 1$mm are typically reported (Optofidelity datasheet). 
Data was stored in a relational database.
Subject details such as handedness, sex and age were recorded along with the associated glyphs. 
Glyphs were stored as a set of associated strokes, corresponding to a period when the subject's finger was in continuous contact with the device panel.  
The coordinate of each touch position was sampled by the touch panel device and this, along with the timestamp of the touch, was stored.
The dataset was reviewed manually and any incorrectly entered glyphs were marked as invalid.
The final dataset contained input from 260 subjects with a total of 20,217 digits gestured and demographic details are summarized in Table~\ref{table:dataset_and_models}a. 

\section{Deep Learning Models}

Two deep learning models were developed. 
One takes an offline glyph bitmap as input and the other takes the polar vectors connecting touch points as input.
The models were implemented using Keras with TensorFlow backend and trained on a NVIDIA TITAN X GPU.

\subsection{Model with Bitmap Input}

The first architecture investigated, as listed in Table \ref{table:dataset_and_models}b, consisted of two convolutional layers and two fully connected layers. Each of the convolutional layers are followed by a rectified linear unit activation layer and a max pooling layer. In the convolutional layers, kernels of size 5x5 were used with a stride of 1. Padding was set to ensure the height and width of the output is the same as the input. The max pooling layers use non-overlapping windows of size 2x2. The result of this is that the output of the second max pooling layer is 7x7. The two fully connected layers come after the aforementioned layers. 50\% dropout is used during training to prevent over fitting and a momentum optimizer, implementing a variation of stochastic gradient descent, was used to minimise the error. The learning rate used for this optimiser was 0.9. Exponential decay was used and the decay rate was set to 0.95. When running for 10 epochs the network took approximately 8 seconds to train on the NVIDIA TITAN X graphics card.

\subsection{Model with Polar Vector Input}

The coordinates of the touch samples were converted to a series of polar vectors.
For each touch point, the vector to the next touch point was calculated. 
The angle of the vector was calculated as the angle to the positive $x$ axis in the range $\pm \pi$ where $+\pi/2$ is vertically upwards.
The length of the vector was expressed in pixels.

The network architecture is listed in Table \ref{table:dataset_and_models}c. 
The input sequences were padded with zeros so that they were they were all the same length as the longest sequence in the dataset, 130 points.
Dropout layers with a dropout rate of 25\% were used to avoid co-adaption of the training data and hence, to reduce overfitting. Max pooling layers with pool size of 2 were used to progressively reduce the number of parameters in the network and hence, reduce the computation required in the training process. In the convolutional layers a kernel size of 5 was used as this was found to capture local features from within the sequence. The activation function used was ReLU as it was found to provide the highest accuracy of the commonly used activation functions. Softmax was used in order to perform the final classification.

Three input cases were considered: angle-only, vector length-only, and both angle and length.
Some of the glyphs include multiple strokes.
Only the longest stroke was input to the network.
This was found to give better accuracy than inputting the entire multi-stroke gesture. Training was considered finished when the validation accuracy did not change for 18 epochs. This typically occurred after 80 epochs. 

\begin{table}[!htb]
    \caption{Dataset and Network Architectures. F\# and P\# refer to the number of features and number of parameters.}
	\label{table:dataset_and_models}
    \begin{minipage}{0.48\textwidth}
    \begin{subtable}{\linewidth}
          \centering
           \caption{Database Demographic}
           \label{table:demographic}
 \begin{tabular}{c c}
 \hline
 Parameter & Number of Entries \\ [0.5ex] 
 \hline
 Male & 126 \\
 Female & 134  \\
 Right Handed & 228 \\
 Left Handed & 32 \\
 Nationalities & 12 \\
 Age Range & 18 - 80 \\
 \hline 
 \end{tabular}
    \end{subtable}
    \begin{subtable}{\linewidth}
      \centering
        \caption{2D Model with Bitmap Input}
        \label{tab:bitmap-model}
        \begin{tabular}{c c c c}
\hline
Layers & Output Size & F \# & P \#\\
\hline
2D Convolution & 28x28 & 32 & 832\\
Max Pooling & 14x14 & - & 0 \\
2D Convolution & 14x14 & 64 & 51264\\
Max Pooling & 7x7 & - & 0 \\
Fully Connected & 512 & - & 1,606,144\\
Dropout & 512 & - & 0\\
Fully Connected & 10 & - & 5130\\
\hline 
\end{tabular}
    \end{subtable}
    \end{minipage}
    \hskip 3ex
        \begin{minipage}{0.45\textwidth}
        \vskip 2ex
    \begin{subtable}{\linewidth}
      \centering
        \caption{1D Model with Polar Vector Input}
        \label{tab:Model-Polar-Input}
        \begin{tabular}{c c c c}
\hline
Layer & Output Size & F \# & P \#\\
\hline
1D Convolution & 126 & 32 & 352 \\
Dropout & 126 & - & 0 \\
1D Convolution & 122 & 32 & 5152 \\
Max Pooling & 61 & - & 0 \\
Dropout & 61 & - & 0 \\
1D Convolution & 57 & 64 & 10304 \\
Max Pooling & 28 & - & 0\\
Dropout & 28 & - & 0 \\
1D Convolution & 28 & 128 & 41088 \\
Max Pooling & 14 & - & 0 \\
Dropout & 14 & - & 0 \\
Flatten & 1792 & - & 0\\
Fully Connected & 128 & - & 229504 \\
Dropout & 128 & - & 0 \\
Fully Connected & 10 & - & 1290 \\
\hline 
\end{tabular}
    \end{subtable} 
        \end{minipage}
\end{table}

\section{Results and Discussion}

The networks were evaluated on the dataset using a 60\% training set, 20\% validation set and 20\% test set split.
The accuracy of the networks is listed in Table \ref{model-comparison}.
It can be seen that the network with bitmap input gives highest accuracy.
The accuracy is close to the results reported in \cite{YannLecun-98} for the NMIST dataset, suggesting that the network is able to cope with the variability of the finger and thumb touch gestures.
In the case of the polar vector input, the best results are obtained by using both angle and distance data.
Also for the polar vector model, using only the longest stroke provided better results than using the full multi-stroke gesture. 
This may be due to a dataset deficiency or the artificial concatenation of the multi-strokes.
The size of the networks is compared in Table \ref{model-comparison}. 
The 2D network is clearly larger due to the number of points on the screen, whereas the 1D network takes only the sequence as input.

\hskip -0.5cm
\begin{minipage}{0.58\textwidth} 
\small
\captionof{table}{Network Accuracy}
\vskip 0.5cm
\label{model-comparison}
\small
\begin{tabular}{ccccc}
   \hline
   Model & Input & Accuracy (\%) & \# of Parameters \\ [0.5ex] 
   \hline
   2D & bitmap & 98.5 & 1,663,370 \\ 
   1D & distance & 76.52 & 287,530 \\ 
   1D & angle & 93.77 & 287,530 \\
   1D & distance \& angle & 95.86 & 287,690 & \\
   \hline 
\end{tabular}
\end{minipage}\hskip 0.1cm
\begin{minipage}{0.4\textwidth}
\includegraphics[width=\textwidth]{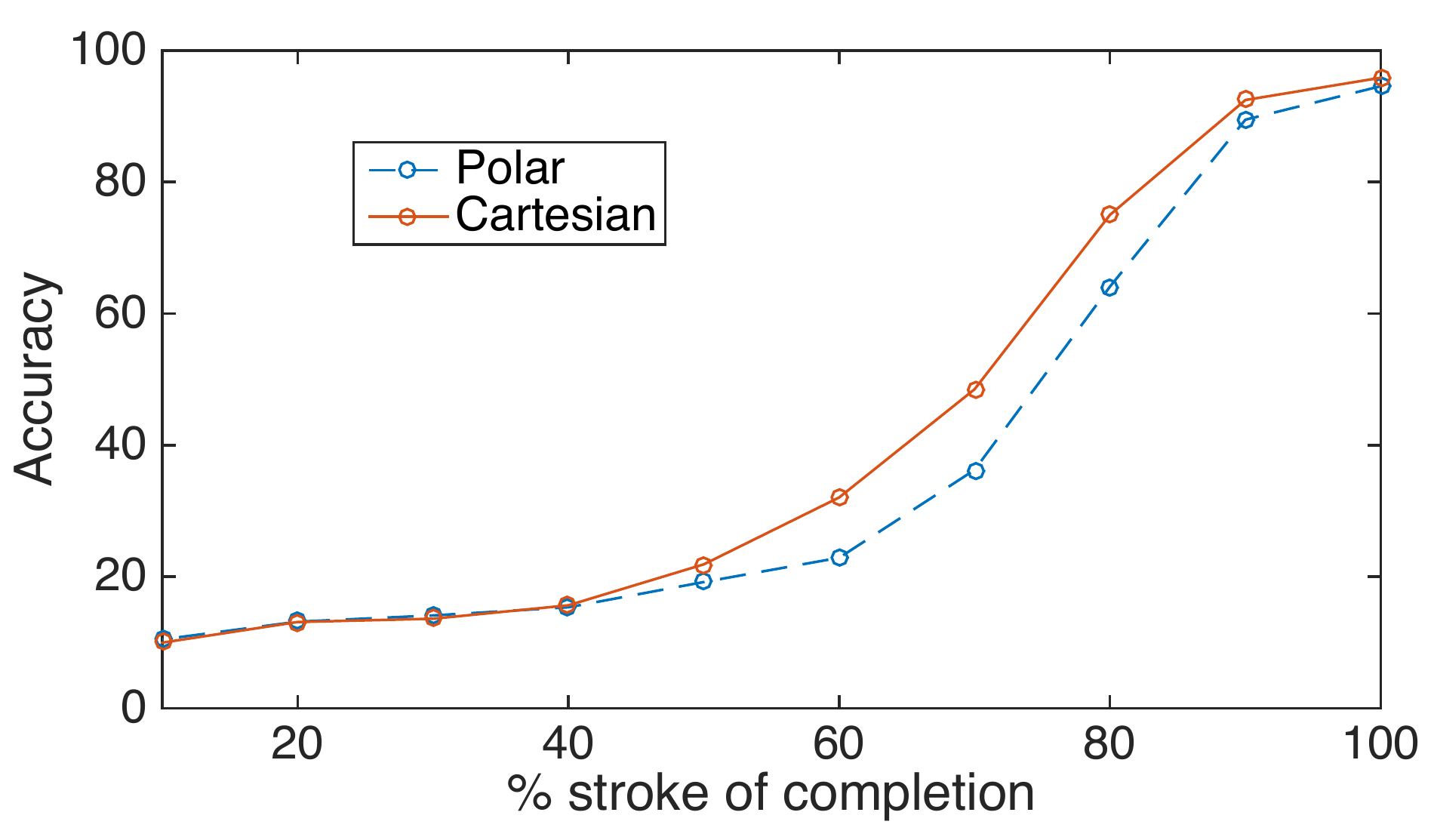}
\small
\vskip -0.3cm
\captionof{figure}{Accuracy vs. stroke completion}
\label{fig:stroke_compeletion}
\end{minipage}


\begin{figure}[!htb]
    \centering
    \label{fig:incorrect-glyphs}
    \begin{minipage}{\textwidth}
    	\begin{minipage}{.18\textwidth}
          \centering
          \includegraphics[width=\textwidth]{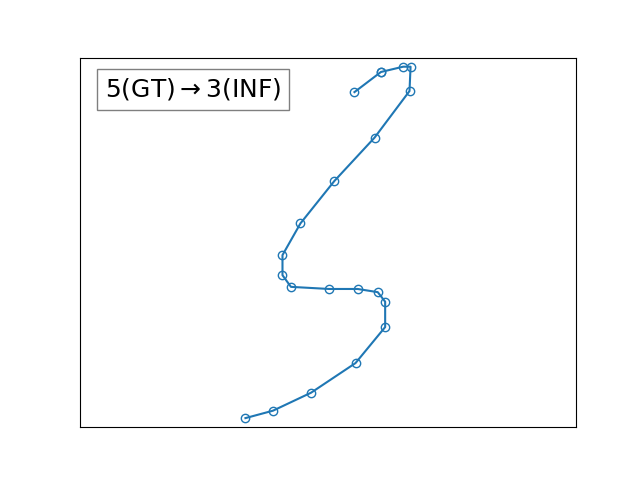}
          \vskip -0.68cm
          \hskip 0.3cm\scriptsize (A)
          \label{fig:glyphA}
    	\end{minipage}
  	\hskip -0.35cm
    	\begin{minipage}{.18\textwidth}
          \centering
          \includegraphics[width=\textwidth]{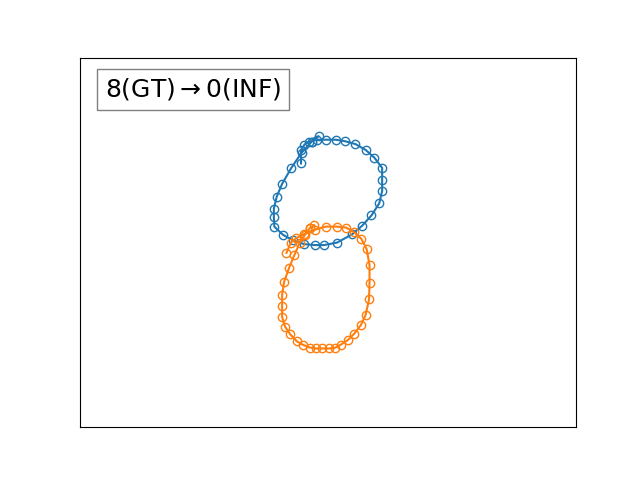}
          \vskip -0.68cm
          \hskip 0.3cm\scriptsize (B)
          \label{fig:glyphB}
    	\end{minipage}
	\hskip -0.35cm
    	\begin{minipage}{.18\textwidth}
          \centering
          \includegraphics[width=\textwidth]{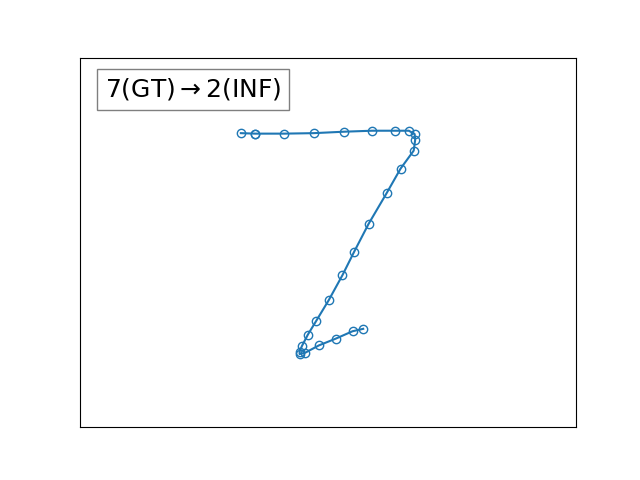}
          \vskip -0.68cm
          \hskip 0.3cm\scriptsize (C)
          \label{fig:glyphC}
    	\end{minipage}
	\hskip -0.35cm
    	\begin{minipage}{.18\textwidth}
          \centering
          \includegraphics[width=\textwidth]{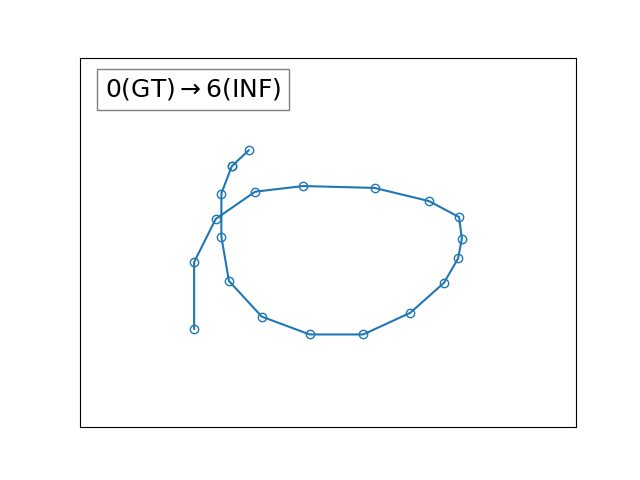}
          \vskip -0.68cm
          \hskip 0.3cm\scriptsize (D)
          \label{fig:glyphD}
    	\end{minipage}
	\hskip -0.35cm
    	\begin{minipage}{.18\textwidth}
          \centering
          \includegraphics[width=\textwidth]{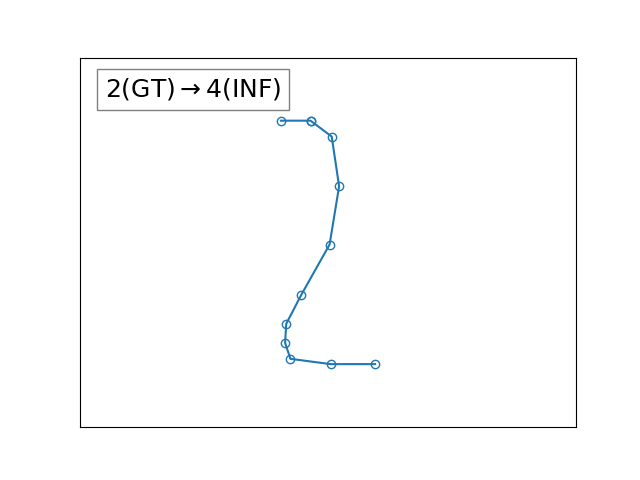}
          \vskip -0.68cm
          \hskip 0.3cm\scriptsize (E)
          \label{fig:glyphE}
    	\end{minipage}
	\hskip -0.35cm
    	\begin{minipage}{.18\textwidth}
          \centering
          \includegraphics[width=\textwidth]{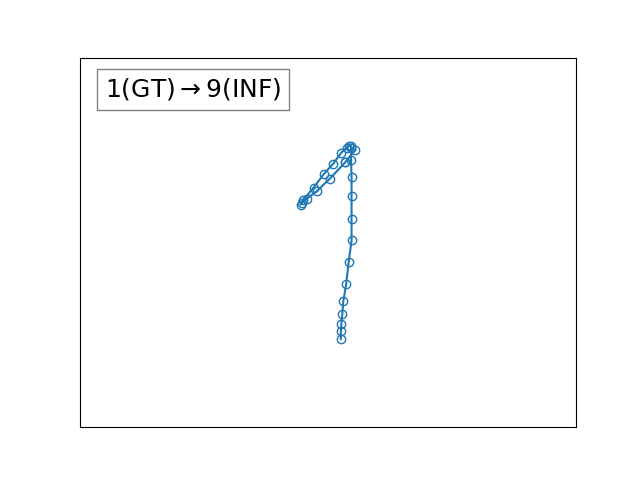}
          \vskip -0.68cm
          \hskip 0.3cm\scriptsize (F)
          \label{fig:glyphF}
    	\end{minipage}
  \end{minipage}
    \caption{ Selection of classification errors. A \& B show glyphs where mis-clasification occurs due to omission of subsequent strokes. C \& D are ambiguous glyphs. E \& F show mis-classification due to glyph formation.
}
\end{figure}

\section{Conclusions and Future Work}

A dataset was created consisting of Arabic numerals recorded on a smartphone touchscreen using single finger or thumb gestures.
Two deep neural networks were trained to recognise the digits.
Both models achieved high accuracy.
One of the models used a novel polar vector data format and had a significantly lower footprint.
In future work, we plan to enhance the accuracy of early digit recognition to accelerate the digit entry process.
It is hoped that the open source dataset described here will facilitate further work on this topic.
The dataset is available at \cite{Gestures}.

\bibliographystyle{apalike}
\bibliography{Bibliography}

\end{document}